\documentclass[letterpaper, 10 pt, conference]{ieeeconf}

\IEEEoverridecommandlockouts
\usepackage{graphics} 
\usepackage{tikz} 
\usepackage{mathptmx} 
\usepackage{amsmath} 
\usepackage{amssymb}  
\usepackage{subfigure}

\usepackage{graphicx}        
\usepackage{multicol}        
\usepackage{multirow}        

\usepackage{bm}
\usepackage{color} 
\usepackage{hyperref} 
\usepackage[utf8]{inputenc}
\usepackage{algpseudocode}
\usepackage{flushend} 
\usepackage{booktabs}
\usepackage[nolist]{acronym}
\usepackage{bbm}

\definecolor{CommentPMN}{rgb}{0.0,0.7,0.0}
\definecolor{CommentMT}{rgb}{0,0.0,0.7}
\definecolor{CommentAB}{rgb}{1.0,0.0,0.0}
\definecolor{CommentSS}{rgb}{0.0,0.4,0.6}

\newcommand{\fix}{\marginpar{FIX}}
\newcommand{\new}{\marginpar{NEW}}

\newcommand{\commentthis}[3]{{{\color{#1} {\textbf{#2} \textbf{#3}}}}}
\newcommand{\ignore}[1]{}

\newcommand{\abnotes}[1]  { \commentthis{CommentAB}{Alex says: }{``#1"}}
\newcommand{\ssnotes}[1]  { \commentthis{CommentSS}{Ștefan says: }{``#1"}}

\ignore{
\renewcommand{\ssnotes}[1]{}
\renewcommand{\abnotes}[1]{}

\renewcommand{\fix}{}
\renewcommand{\new}{}
}

\newcommand{\eg}{\emph{e.g.}}
\newcommand{\ie}{\emph{i.e.}}
\newcommand{\etal}{\emph{et al.}}

\graphicspath{{figures/}}

\title{Meshed Up: Learnt Error Correction in 3D Reconstructions}

\author{Michael Tanner$^\dagger\star$ \and Ștefan Săftescu$^\dagger\star$ \and Alex Bewley$^\dagger$ \and Paul Newman$^\dagger$
  \thanks{$^\dagger$Oxford Robotics Institute, University of Oxford, United Kingdom; \newline 
  {\tt\small mtanner,stefan,bewley,pnewman@robots.ox.ac.uk}}}%

\newcommand{\borg}{BOR$^2$G}

\begin{acronym}
  \acro{\borg{}}{Building Optimal Regularised Reconstructions with GPUs}
  \acro{TV}{Total Variation}
  \acro{TGV}{Total Generalised Variation}
  \acro{HVG}{Hashing Voxel Grid}
  \acro{KCDE}{Kernel Conditional Density Estimation}
  \acro{1D}{one dimensional}
  \acro{2D}{two dimensional}
  \acro{3D}{three dimensional}
  \acro{SDF}{Signed Distance Function}
  \acro{TSDF}{Truncated Signed Distance Function}
  \acro{MAP}{\emph{maximum a posteriori}}
  \acro{VO}{Visual Odometry}
  \acro{IMU}{Inertial Measurement Unit}
  \acro{SLAM}{Simultaneous Localisation and Mapping}
  \acro{DTAM}{Dense Tracking and Mapping}
  \acro{CUDA}{Compute Unified Device Architecture}
  \acro{PDF}{Probability Density Function}
  \acro{GPGPU}{General Purpose Graphics Processing Unit}
  \acro{GPU}{Graphics Processing Unit}
  \acrodefplural{GPU}[GPUs]{Graphics Processing Units}
  \acro{CPU}{Central Processing Unit}
  \acro{SfM}{Structure from Motion}
  \acro{ORI}{Oxford Robotics Institute}
  \acro{NASA}{National Aeronautics and Space Administration}
  \acro{LIBELAS}{Library for Efficient Large-scale Stereo Matching}
  \acro{LUT}{lookup table}
  \acrodefplural{LUT}[LUTs]{lookup tables}
  \acro{LF}{Legendre-Fenchel}
  \acro{GLSL}{OpenGL Shading Language}
  \acro{RMSE}{root mean square error}
  \acro{CNN}{convolutional neural network}
  \acro{KITTI-VO}{KITTI visual odometry}
  
\end{acronym}

\begin{document}

\maketitle

\begin{abstract}

  Dense reconstructions often contain errors that prior work has so far minimised using high quality sensors and regularising the output. Nevertheless, errors still persist. 
  This paper proposes a machine learning technique to identify errors in \ac{3D} meshes. 
  Beyond simply identifying errors, our method quantifies both the magnitude and the direction of depth estimate errors when viewing the scene. This enables us to improve the reconstruction accuracy. 
  
  We train a suitably deep network architecture with two \ac{3D} meshes: a high-quality laser reconstruction, and a lower quality stereo image reconstruction. The network predicts the amount of error in the lower quality reconstruction with respect to the high-quality one, having only view the former through its input. We evaluate our approach by correcting \ac{2D} inverse-depth images extracted from the 3D model, and show that our method improves the quality of these depth reconstructions by up to a relative 10\% RMSE.
\end{abstract}

\section{Introduction}



This paper is about detecting and rectifying mess ups in dense reconstructions. Dense reconstruction as a mapping from input images to 3D meshes is a well studied area. However much of that work is feed-forward in the sense that it strives to produce the best mesh in an open loop. In this work we consider how, given an arbitrary generated mesh and the images that were used to create it, one might correct reconstruction errors post-hoc. 

\begin{figure}[t!]
  \centering{
    \subfigure[Camera-only Reconstruction]{\includegraphics[width=1.0\columnwidth]{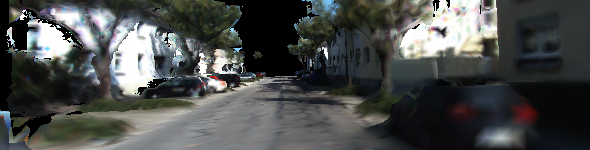}}  
    \subfigure[Inverse-depth image]{\includegraphics[width=1.0\columnwidth]{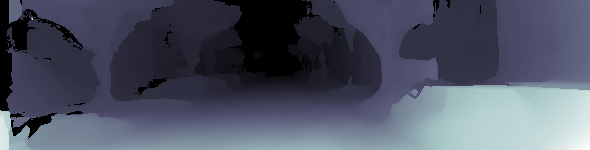}}  
    \subfigure[Predicted amount of error]{\includegraphics[width=1.0\columnwidth]{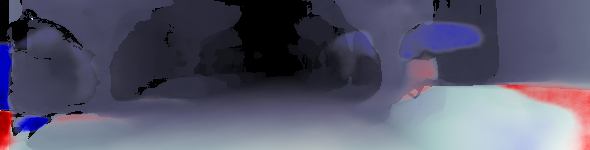}}  
    \subfigure[Corrected inverse-depth image]{\includegraphics[width=1.0\columnwidth]{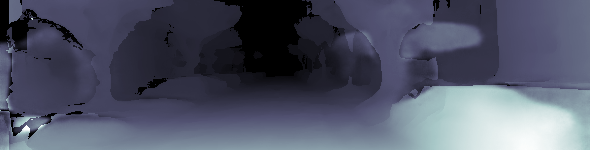}}  
  }
  \caption{An illustration of using the \ac{CNN} presented in this paper to correct an inverse-depth image generated from a dense \ac{3D} reconstruction. Structural errors in the reconstruction that are not obvious in (a) due to the lack of colours, are evident in (b), especially around the car. An overlay of the error predicted by the CNN’s over (b) is displayed in (c). Note that this predicted error is signed, with positive error shown in red and negative error shown in blue. (d) shows how simply subtracting this error from the inverse depth image results in a better representation of the scene (\eg{} the car edges are more defined).}
  \label{fig:headline}
  \vspace{-4ex}
\end{figure}

Prior efforts towards improving the accuracy of generated \ac{3D} meshes often focused on regularising the camera depth maps \cite{newcombe2011dtam}, using expensive high-quality sensors \cite{bok2014sensor}, or regularising the dense-reconstruction output \cite{wright2017algorithmic}. While regularisation based approaches impose structure to smooth local surfaces, gross reconstruction errors still remain. In contrast the focus of this work is to identify error prone areas in \ac{3D} reconstructions. The goal is to facilitate the removal or repair of these mesh regions in order to improve overall mapping accuracy. As an example of what we will describe Figure~\ref{fig:headline} illustrates how the most inaccurate areas in a reconstruction are identified and corrected in a \ac{2D} representation of the scene.

\begin{figure*}[t!]
  \centering{
    
    \subfigure[Camera-only Reconstruction (Colour)]{\includegraphics[width=1.0\columnwidth]{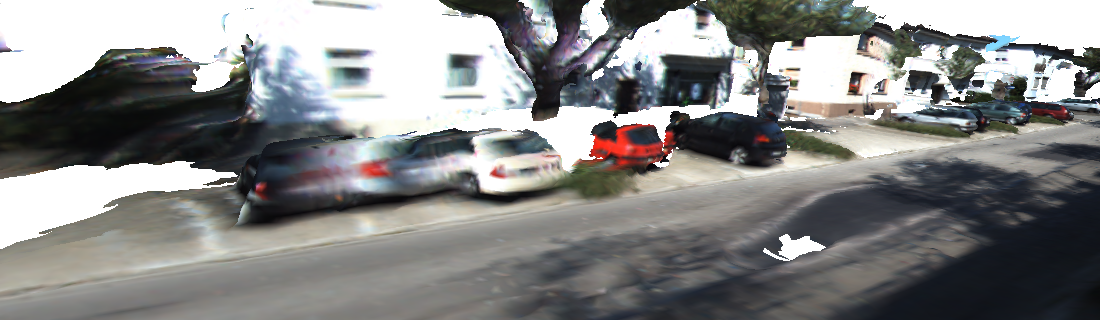}}
    \subfigure[Camera-only Reconstruction (Mesh)]{  \includegraphics[width=1.0\columnwidth]{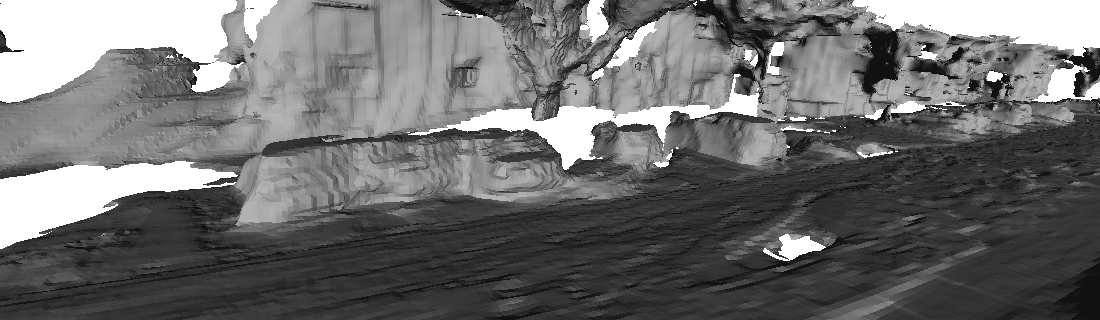}}
    \subfigure[Laser Reconstruction (Colour)]{    \includegraphics[width=1.0\columnwidth]{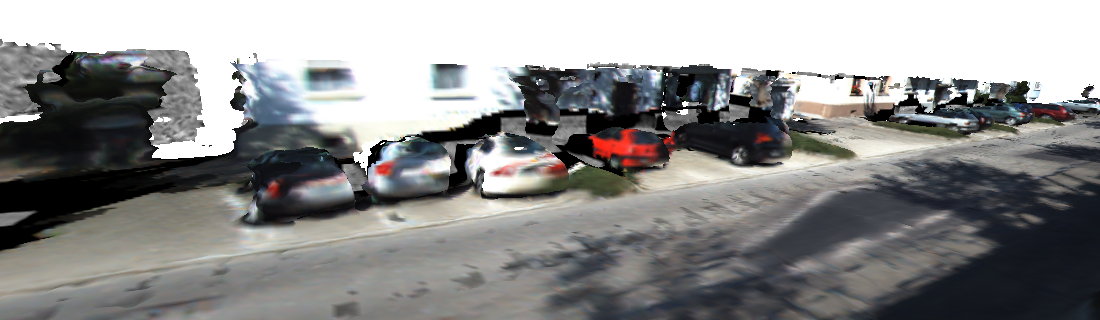}}
    \subfigure[Laser Reconstruction (Mesh)]{      \includegraphics[width=1.0\columnwidth]{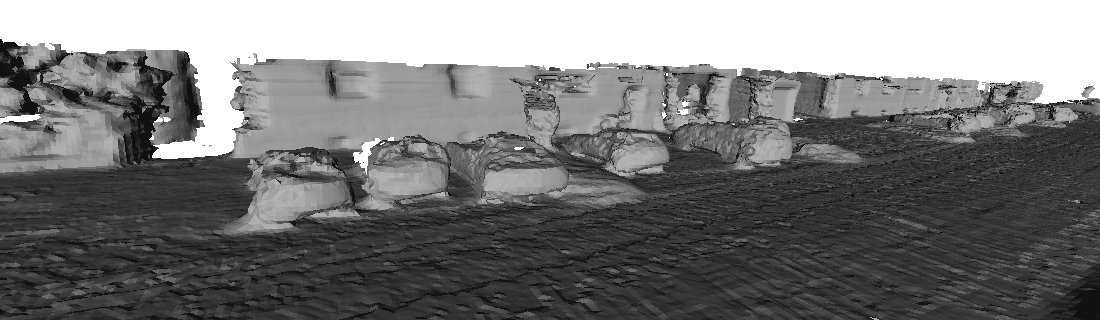}}
  }
  \caption{
  Camera depth-map and laser mesh reconstructions generated using \borg{}~\cite{tanner2016bor}, used for network inputs and ground truth respectively.
  In the top examples ((a) and (b)) one can easily recognise the ``smearing'' between the cars and the holes in the road are incorrect.
  The goal of this work is to use a neural network to automatically recognise these regions in depth-map reconstructions by leveraging training data derived from laser reconstructions (\eg{}, (c) and (d)).
  }
  \label{fig:eg_image_vs_laser}
  \vspace{-4ex}
\end{figure*}

To realise this capability, a mapping is learnt from rasterised geometric and appearance features to the error in the camera-based reconstruction. Accurate laser reconstructions serve as our ground-truth to train a \ac{CNN} that estimates the quality of depth-map reconstructions.
A simple example of this process is shown in Figure~\ref{fig:eg_image_vs_laser}, which displays the same scene reconstructed by laser and camera input data.
Compared to the laser reconstruction's colour and mesh images, the regions in the camera reconstruction that are likely incorrect can readily be identified -- \eg{} holes in the road and the ``smearing'' area between the cars.
Ideally, we would have a similar but automated way of perceiving the mesh outputs to identify erroneous regions.
In this paper, we present our method for training a neural network to recognise and highlight these regions.



\begin{figure*}[t!]
\centering{
    \includegraphics[width=1.7\columnwidth]{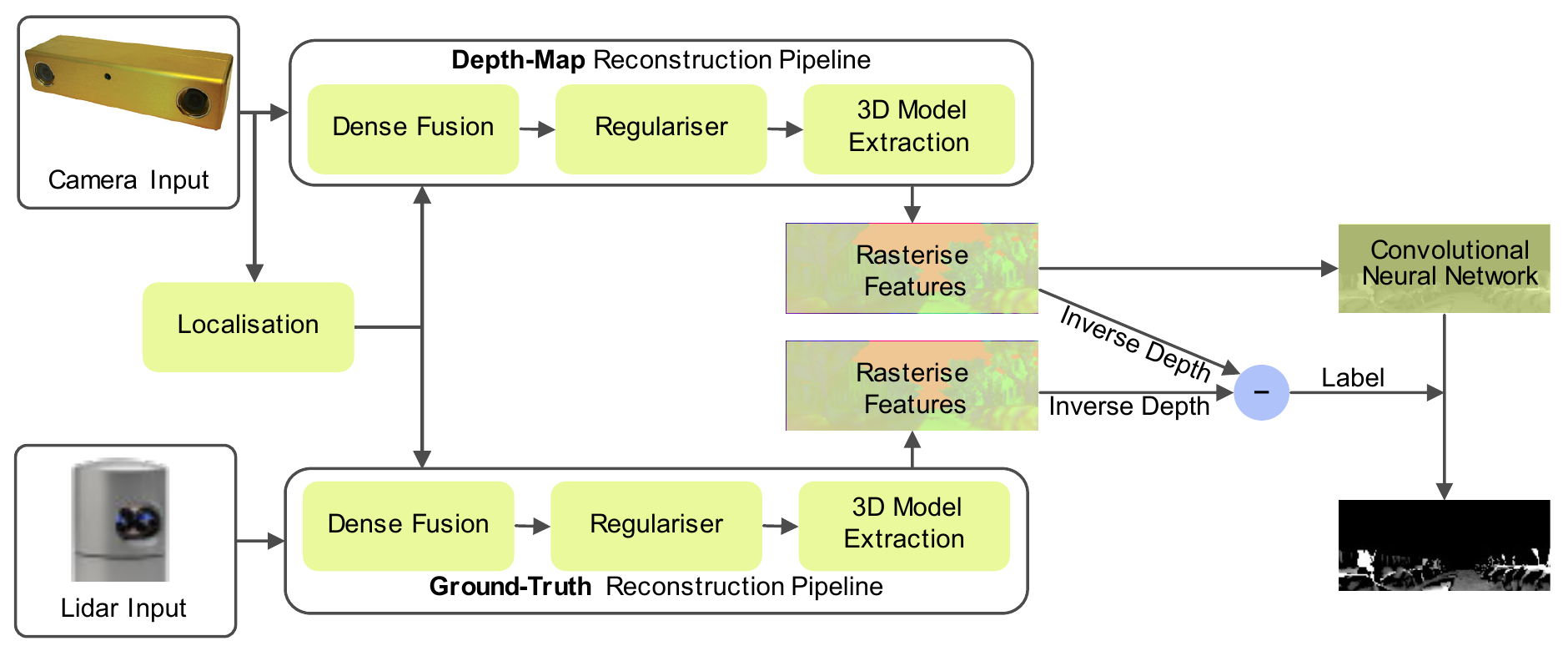}
\caption[Machine Learning Data-Flow Pipeline]{
  Machine Learning Data-Flow Pipeline.
  To train our network, we create separate depth-map and laser dense reconstructions (using the techniques discussed in \cite{tanner_denser_2016}).
  We create ground-truth data by subtracting the rasterised inverse-depth images from each reconstruction.
  The neural network trains on rasterised feature images (see Figure~\ref{fig:features}) to learn to generalise the ground-truth error in new scenes.
}
\label{fig:pipeline}
\vspace{-3ex}
}
\end{figure*}

A high-level overview of the proposed system is presented in Figure~\ref{fig:pipeline} that shows the framework for building the meshes used in training a \ac{CNN}. Two dense reconstructions are used: one constructed using camera-only depth-maps and another constructed using more accurate laser data.
Both geometric and appearance features are extracted from each reconstruction and rasterised to a \ac{2D} image suitable for training the \ac{CNN} (Section \ref{sec:mesh_features}).

In Section~\ref{sec:architecture}, we describe our network architecture and loss function based on related work in the area of monocular depth estimation \cite{laina_deeper_2016} and re-purpose it towards the task of correcting for mesh errors.
In particular, we investigate the potential and value of providing a diverse set of features (\eg{} colour, inverse depth, triangle edge ratios and surface normals) to the network which are not commonly available without a mesh representation.
Finally, in Section \ref{sec:experiments} we demonstrate the efficacy of the proposed system in identifying and correcting errors on three meshes created using sequences from the \ac{KITTI-VO} \cite{geiger2012kitti-vo} data set.
Within this framework, the following contributions are made:
\begin{enumerate}
    \item The novel use of a \ac{CNN} applied to rasterised mesh features is shown to identify error-prone regions in a low quality mesh \ac{3D} reconstruction;
    \item Using the network predictions to correct for error in the inverse-depth images leads to significant improvements over what is extracted from the original mesh;
    \item We analyse which of the rasterised mesh features are most informative to the \ac{CNN} through an ablation study on the trained network.
\end{enumerate}



\section{Mesh Features}\label{sec:mesh_features}
Much work in the deep learning literature relies upon simple RGB images as the input \cite{eigen_depth_2014}\cite{liu_deep_2015}\cite{laina_deeper_2016}.
However, as our problem formulation assumes that a dense \ac{3D} model is available, we can extract a much richer set of features from the reconstruction. 
Throughout this work we use the dense mesh reconstruction proposed by Tanner \etal{}~\cite{tanner2016bor} referred to as \borg{}, although our framework is agnostic to the \ac{3D} mesh pipeline.

\subsection{Feature Creation}\label{sec:features}

\begin{figure}[t!]
  \centering{
    \subfigure[Image from the KITTI-VO Data Set]{      \includegraphics[width=0.8\columnwidth]{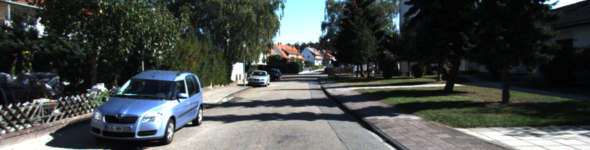}}
    \subfigure[Colour Reconstruction]{      \includegraphics[width=0.8\columnwidth]{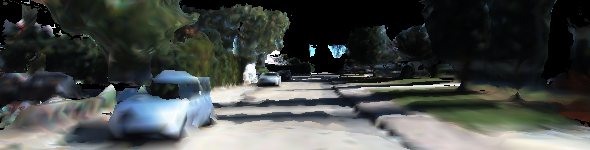}}
    \subfigure[Inverse-depth]{            \includegraphics[width=0.8\columnwidth]{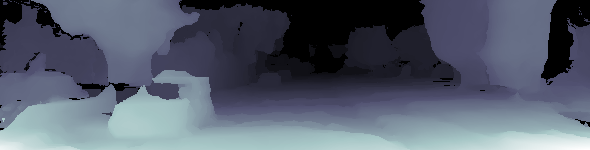}}
    \subfigure[Triangle Area]{      \includegraphics[width=0.8\columnwidth]{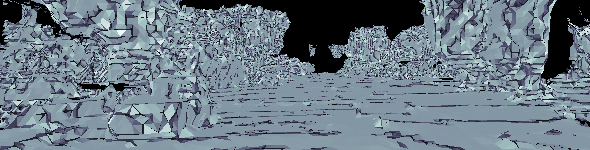}}
    \subfigure[Triangle Surface Normal]{    \includegraphics[width=0.8\columnwidth]{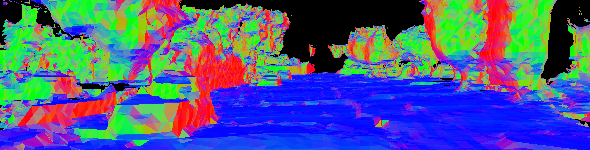}}
    \subfigure[Triangle Edge Length Ratios]{\includegraphics[width=0.8\columnwidth]{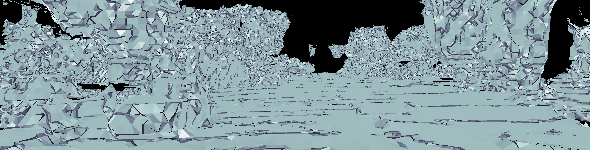}}
    \subfigure[Surface to Camera Angle]{\includegraphics[width=0.8\columnwidth]{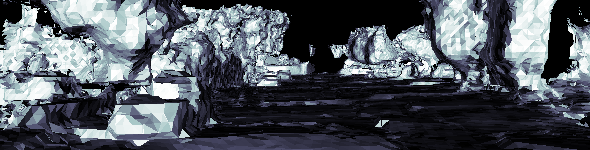}}
  }
  \caption[Example Input Features]{
  Example Input Features.
  As our dense reconstructions provide a \ac{3D} model of our operating environment, we can extract a variety of features.
  Pictured above are example features our \ac{GLSL} pipeline currently extracts. The network has the potential to use these low level features in its intermediate representation.
  }
  \label{fig:features}
  \vspace{-5mm}
\end{figure}

To extract dense 2D features suitable for a \ac{CNN}, we begin by using \ac{GLSL} to ``fly'' a virtual camera through the \ac{3D} models.
Our \texttt{vertex} shader transforms the \ac{3D} model's vertices into the camera reference frame to compute the depth of each vertex and the normalised camera vector.
These camera-frame points are then grouped into triangles and passed along to a \texttt{geometry} shader that computes the triangle's area, surface normal, and edge lengths.
Finally, the \texttt{fragment} shader processes each of the preceding feature values and rasterises them into individual pixels in feature images, as shown in Figure~\ref{fig:features}.
This \ac{GLSL} process enables us to extract six feature images:  two per-pixel features (RGB and depth) and four per-triangle features (area, surface normal, edge length ratio, and surface-to-camera angle).
Intuitively, the mesh-triangle-derived features contain a lot of information; they provide geometric cues that have been accumulated from multiple depth-maps and integrated into a \ac{3D} reconstruction by \borg{}. We look at whether this intuition is verified in Section~\ref{sec:experiments}.

\subsection{Ground-Truth Data}\label{sec:gt}


For a ground-truth reference, we construct a mesh using the same reconstruction pipeline as above. The same virtual camera trajectory is used to create a sequence of feature images from both the depth-map and laser reconstructions. Since the goal is to train a neural network to recognise \emph{errors} in the depth-map reconstruction, we must first compute \emph{ground-truth errors} with which to train the network. Rather than directly subtracting depths, we instead elect to use inverse-depth. This way, the network is encouraged to emphasise errors in foreground objects – which are more likely to have complex geometry compared to background surfaces.
Specifically, we compute the per-pixel ground-truth ($\Delta_{gt}$) as,
\begin{equation}\label{eq:gt}
  \Delta_{gt} = \frac{A}{d_c} - \frac{A}{d_l},
\end{equation}%
where $d_c$ and $d_l$ are pixels in the depth-map features for the camera and laser reconstruction, respectively, with $A$ being the scaling constant.
In disparity images from stereo cameras, $A = f_x b$ where $f_x$ is the camera's $x$ focal length and $b$ is the baseline distance between the left and right image sensors. Because our approach is agnostic of the input sensor, $A$ can be arbitrary, and we set $A = 1$ when evaluating performance (Section~\ref{sec:experiments}).

\section{Reconstruction Error Prediction }\label{sec:architecture}


\subsection{Network Architecture}
For our network architecture, we based our network on Lania \emph{et al}.'s Fully Convolutional Residual Network \cite{laina_deeper_2016} with concatenated ReLU activation functions on the intermediate layers \cite{Shang2016}.
This network was designed to infer per-pixel depth from a single RGB image, a task highly correlated with our aim to compute estimated depth \emph{error} given a set of feature inputs. In this work, the input layer is generalised to accept $F$ input channels dependant on the number of active features used from the previous section.
Table \ref{tab:network-architecture} provides high-level details of the \ac{CNN} where a series of residual blocks based on the \emph{ResNet-50} architecture \cite{He2016} is followed by up-projection blocks, proposed in \cite{laina_deeper_2016}. Note the fractional strides denote spatial upsampling. The final layer is a simple $3 \times 3$ convolutional layer which outputs real-valued estimates corresponding to the inverse-depth error at each pixel location.

\begin{table}[t!]
    \centering
    \caption{Overview of the CNN architecture for error prediction}
    \begin{tabular}{p{0.40\linewidth}cc}
        \toprule
        \textbf{Block Type} & \textbf{Filter Size/Stride} & \textbf{Output Size} \\
        \midrule
        Input & - & $64\times 96\times F$ \\
        Convolution & $7\times 7$/$2$ & $32\times 48\times 64$ \\
        Max Pool & $3\times 3$/$2$ & $16\times 24\times 64$ \\
        Residual, Residual, Projection & $3\times 3$/$2$ & $16\times 24\times 256$ \\
        Residual, Residual, Projection & $3\times 3$/$2$ & $8\times 12\times 512$ \\
        Residual, Residual, Projection & $3\times 3$/$2$ & $4\times 6\times 1024$ \\
        Residual, Residual & $3\times 3$/$1$ & $2\times 3\times 2048$ \\
        Up-projection & $3\times 3$/$\frac{1}{2}$ & $4\times 6\times 1024$ \\
        Up-projection & $3\times 3$/$\frac{1}{2}$ & $8\times 12\times 512$ \\
        Up-projection & $3\times 3$/$\frac{1}{2}$ & $16\times 24\times 256$ \\
        Up-projection & $3\times 3$/$\frac{1}{2}$ & $32\times 48\times 128$ \\
        Up-projection & $3\times 3$/$\frac{1}{2}$ & $64\times 96\times 32$ \\
        Convolution & $3\times 3$/$1$ & $64\times 96\times 1$ \\
        \bottomrule
    \end{tabular}
    \vspace{-0.5cm}
\label{tab:network-architecture}
\end{table}

\subsection{Generalisation Capacity}

To be able to deploy a learnt \ac{CNN} to new meshes created without a high-fidelity laser we need our network generalise to unseen data. To this end, three techniques are employed: cropping, downsampling, and regularisation. 
Firstly, we randomly perturb and crop all input feature images before providing them to the network. After each epoch of training (\ie{} the network has viewed all the training images), the next epoch will receive a slightly different cropped region of each input image.  This prevents the network from associating a specific pixel location in the training data with its corresponding output.

Secondly, the feature maps are gradually downsampled via projection blocks (from \emph{ResNet-50}) creating a bottleneck, thus reducing representational capacity.  This also provides greater context to the convolutional filters in the later layers of the network by increasing the size of their receptive field. 

Thirdly, we implement a $L_2$ weight regulariser to further constrain the representational capacity by preventing the network from over-relying on the cost function at the expense of generalisation performance.

\subsection{Loss Function}

Several loss functions are widely used in machine learning applications.
The $L_2$ norm is traditionally popular because it heavily penalises large errors and is smooth.
However, unlike the $L_1$ norm, it has a near zero gradient for small errors, thus is often unable to drive the error completely to zero.
Over the years, researchers have proposed alternative norms which combine the ``best'' (based on application) characteristics of each of these norms.
The Huber norm uses $L_2$ near the origin and $L_1$ elsewhere, while BerHu uses the $L_1$ norm near the origin and the $L_2$ elsewhere \cite{owen_robust_2007}.
We choose BerHu as our cost function on the networks error prediction since the $L_1$ term places a higher penalty on small errors (when compared to Huber or $L_2$) while still providing strong penalties for larger errors. This is in contrast to regularising depth maps where the Huber norm is more favourable to capture high gradient edges around objects \cite{newcombe2011dtam}. These benefits were also observed in the related task of depth prediction \cite{laina_deeper_2016}.

\section{Experiments}\label{sec:experiments}


\subsection{Experimental Setup}
\paragraph*{Dataset}
Three sequences from the \ac{KITTI-VO} dataset where used as the input to the reconstruction pipeline.
The input of the network consists of features from the camera based mesh, while the ground truth used for both training and evaluation comes from the laser based mesh, as described in Section~\ref{sec:gt}.
For all experiments, the KITTI-VO sequences 00, 05 and 06 were used for training and leave-one-out style cross-validation.
All three sequences are predominantly in urban environments with small amounts of visible vegetation.
Using \ac{GLSL}, we created a virtual camera to project the dense reconstruction into input feature images. The trajectory of the virtual camera follows the original trajectory in the KITTI-VO sequences. This enables the use of real colour images as input alongside the generated features. Both the camera and laser reconstruction meshes were generated with a fixed voxel width of 0.2\,m.

\paragraph*{Network Training and Inference}
The network module was implemented in Python using the TensorFlow framework. 
The network weights were optimised using the ADAM solver \cite{kingma2014adam} with a learning rate of $10^{-4}$ for 250 epochs over two of the sequences. At this point, the value of the loss function appears converged with variance coming form the stochasticity of the mini-batch sampling.
To reduce this variance and confirm convergence, the optimisation is run a further 50 epochs with the learning rate reduced to $10^{-5}$.
Other training hyper-parameters include a batch size of 16 and a $L_2$ weight decay factor of $10^{-6}$ per training step. Inference takes, on average, 72\,ms per half-size KITTI image and aassociated features.

\paragraph*{Performance Metrics}
This work uses two different metrics to measure the resulting performance.
The first metric is the commonly used \ac{RMSE} which provides a quantitative measure of per pixel error, computed as follows:
\begin{equation*}
\textrm{RMSE} = \sqrt{\frac{1}{n}\sum_{X}{(d^* - d_l)^2}},
\end{equation*}
where $d_l$ is the ground truth depth, $d^*$ is the predicted depth, $X$ is the set of valid pixels, $n$ is the cardinality of $X$. 

Our second metric measures the accuracy of our network’s ability to estimate errors under a given threshold, serving as an indication of how often our estimate is correct.
The thresholded accuracy measure from \cite{Liu2015a} is essentially the expectation that a given pixel in $X$ is less than a threashold $thr^k$:
\begin{equation*}\label{eq:thresholded_accuracy}
\delta_k = \mathbb{E}_X\left[\mathbb{I}\left(\textrm{max}\left(\frac{d_l}{d^*}, \frac{d^*}{d_l}\right) < thr^k\right)\right],
\end{equation*}
where $\mathbb{I(\cdot)}$ represents the indicator function.
As in \cite{Liu2015a}, ~$thr = 1.25$, and $k \in \{1, 2, 3\}$. 

\subsection{Correcting Depth Maps}
\begin{figure*}[]
  \centering{
    \subfigure[Original Image]{\includegraphics[width=1.0\columnwidth]{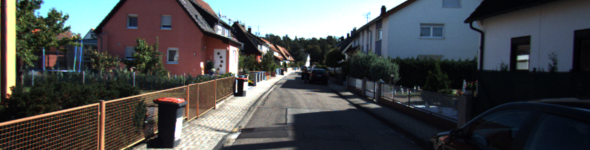}}
    \subfigure[Inverse-depth Image of Camera Reconstruction]{\includegraphics[width=1.0\columnwidth]{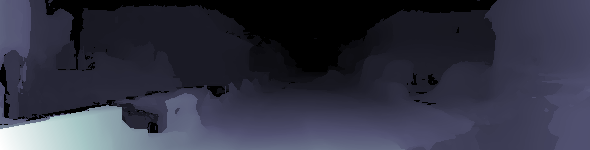}}
    \subfigure[\ac{CNN} Predicted Error]{\includegraphics[width=1.0\columnwidth]{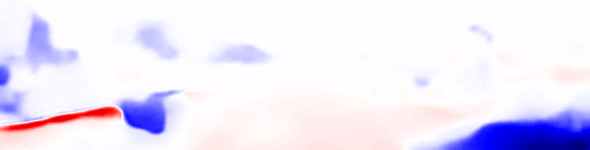}}
    \subfigure[Corrected Inverse-depth Image]{\includegraphics[width=1.0\columnwidth]{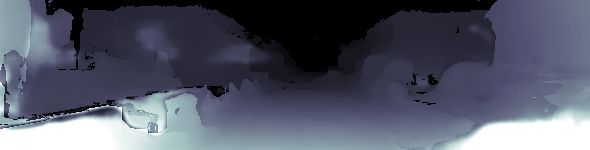}}
    \subfigure[Ground-Truth Error]{\includegraphics[width=1.0\columnwidth]{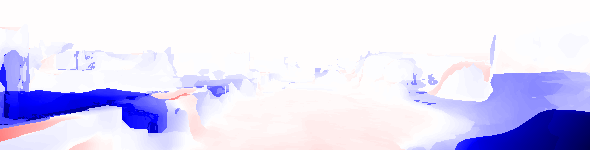}}
    \subfigure[Inverse-depth Image of Laser Reconstruction]{\includegraphics[width=1.0\columnwidth]{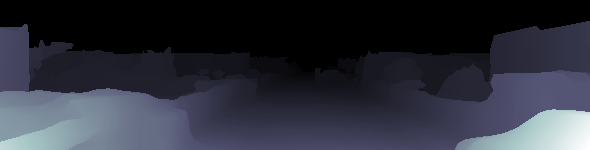}}
  }
  \caption[Example Network Prediction]{
    An example demonstrating how our network corrects for reconstruction error on a test set image. (a), (b): A scene where gross errors in the fence on the left and car on the right are captured within the output predictions of the network (c). Here the colour intensity shows the magnitude of the errors with red indicating positive (camera depth is too close) and blue for negative (camera depth is too far) depth errors. 
    Trivially applying the output of out network as a correction to the low quality inverse depth map (b) produces a higher quality inverse depth estimate (d), capturing structure missed in the original camera based reconstruction.
    
  }
  \label{fig:inout1}
  \vspace{-4mm}
\end{figure*}

The central question of this paper is whether we can correct 3D mesh reconstructions using a \ac{CNN}. 
As a proxy, the mesh surface projected as an inverse-depth image and used to represent the local 3D scene. 
For our baseline reference, we compare the inverse-depth image from the camera-only reconstruction (created using \borg{}) with the equivalent inverse-depth from the laser-only reconstruction, thus $d^* = d_c$. 
To obtain the refined depth-map, we use the network output $\Delta^*$ to correct the depth-map reconstruction, giving us $d^* = d_c - 1/\Delta^*$. An example of this process is shown in Figure \ref{fig:inout1}.

\begin{figure}[]
  \centering
  \includegraphics[width=0.95\columnwidth]{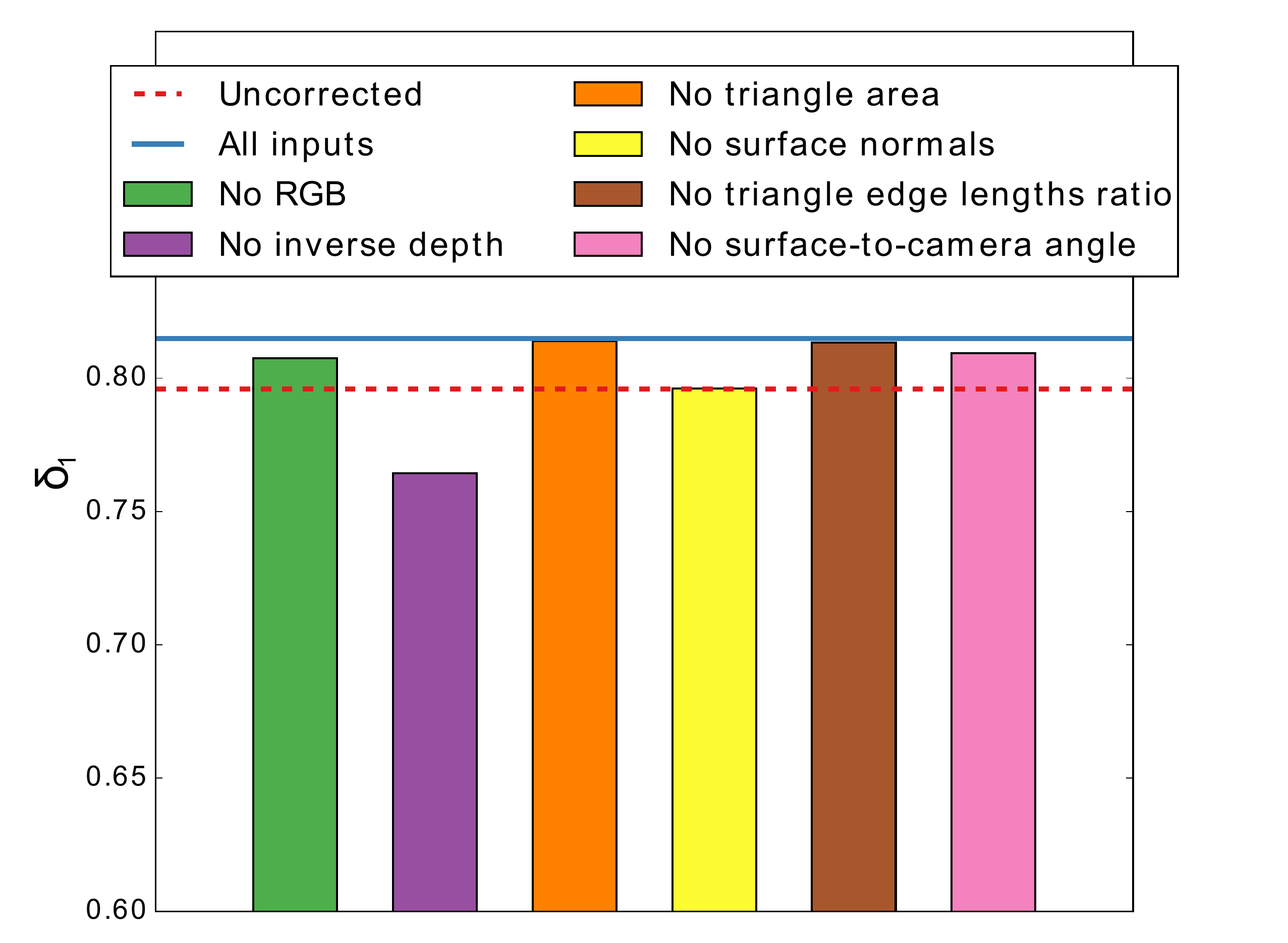}
  \caption{Analysis of our framework's sensitivity to disabling different input features on the depth correction accuracy $\delta_1$.}
  \label{fig:ablation}
  \vspace{-4mm}
\end{figure}

\begin{figure*}[t]
  \centering{
    \subfigure[Colour Reconstruction]{%
    \begin{minipage}{0.67\columnwidth}
    \includegraphics[width=\textwidth]{images/02239-image_rgb}\\[1.5ex]
    \includegraphics[width=\textwidth]{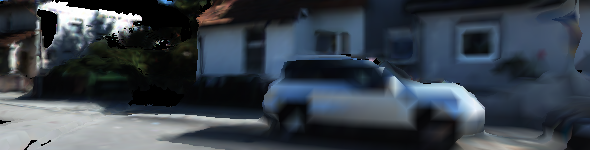}\\[1.5ex]
    \includegraphics[width=\textwidth]{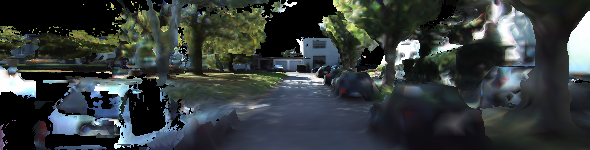}\\[1.5ex]
    \includegraphics[width=\textwidth]{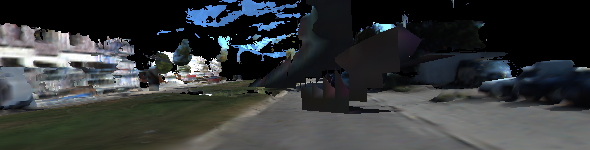}\\
    \end{minipage}
    }%
    \subfigure[Ground-Truth Error]{%
    \begin{minipage}{0.67\columnwidth}
    \includegraphics[width=\textwidth]{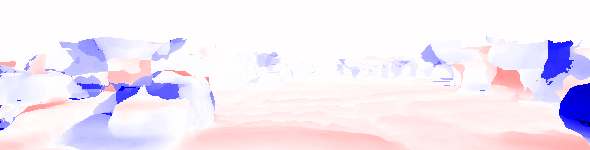}\\[1.5ex]
    \includegraphics[width=\textwidth]{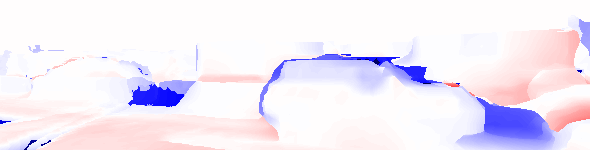}\\[1.5ex]
    \includegraphics[width=\textwidth]{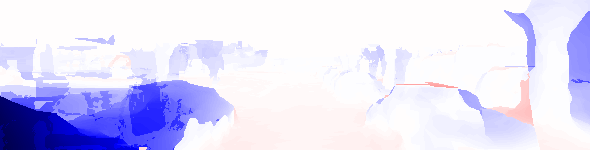}\\[1.5ex]
    \includegraphics[width=\textwidth]{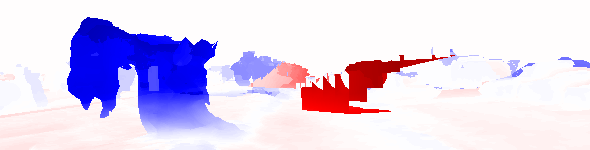}\\
    \end{minipage}
    }%
    \subfigure[CNN Predicted Error]{%
    \begin{minipage}{0.67\columnwidth}
    \includegraphics[width=\textwidth]{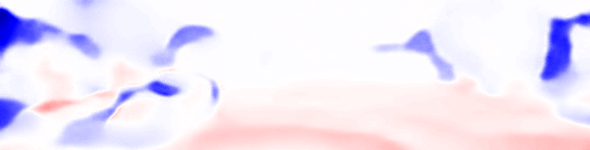}\\[1.5ex]
    \includegraphics[width=\textwidth]{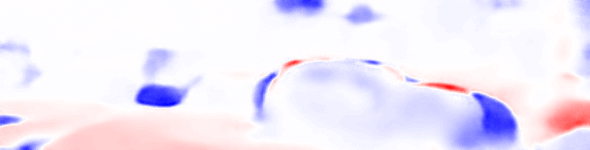}\\[1.5ex]
    \includegraphics[width=\textwidth]{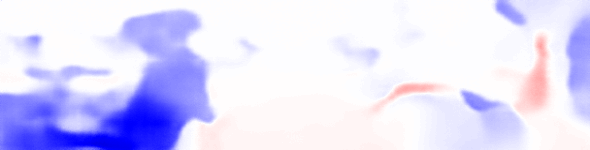}\\[1.5ex]
    \includegraphics[width=\textwidth]{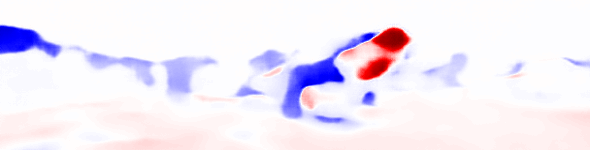}\\
    \end{minipage}
    }%
  }
  \caption[Example Ground-Truth vs. Network Prediction]{
  Example Ground-Truth and Network Prediction.
  Given the scene present in (a), our network successfully recognises (c) the most error-prone regions of the reconstruction -- \eg{} unlikely borders around cars, fences, and buildings. The last row shows an instance of a failure where the grass on the left of the road appears to be consistent from the camera view point and the car ahead is dynamic and therefore significantly challenges the static world assumption in the reconstruction process.
  }
  \label{fig:gt_vs_pred}
  \vspace{-4ex}
\end{figure*}

Quantitatively, we find our CNN based correction provides a 10\% relative \ac{RMSE} improvement over the baseline generated by \borg{}. Furthermore, it is consistently more accurate over different thresholds, indicating better depth values both across more pixels.
A qualitative visualisation of the network output is provided in Figure \ref{fig:gt_vs_pred}, along side the ground truth difference in camera to laser inverse depth maps. This visualisation shows that the network is capable of identifying small errors over broad regions (\eg the red offset in the road level potentially caused by imperfect calibration) as well as large but narrow errors commonly found at object boundaries.

\subsection{Feature Ablation Study}

In the previous experiment, all features were provided to the CNN under the assumption that some features may contain more information than others and that the network can choose to ignore certain features if they contain no information.
This will be reflected by measuring the sensitivity in the network output when a specific feature is withheld from the input.
It is also possible that the full set of features contains redundant information which would result in the network spreading its internal importance weighting across multiple features. Such behaviour is implicitly encouraged through the $L_2$ weight regularisation placed on the network, meaning that small weights across multiple channels are favoured over large weights on single channels. If the network learns to use redundant features then it should be more robust to the absence of one of the features if the equivalent information could be found elsewhere.

Figure~\ref{fig:ablation} shows the relative feature importance evaluated by disabling one feature at a time and monitoring the influence on the inverse depth accuracy $\delta_1$. 
Crucially, the inverse-depth from the camera reconstruction was considered to be the most important.
On the other hand, mesh specific features such as triangle area and edge length ratio (Figure~\ref{fig:features} (d)(f)) appear to be least useful.
To test if redundancy is explaining this lack of sensitivity, we keep only RGB, inverse-depth and normals, to fine-tune the network (Table \ref{tab:ablation_summary}). Here, the CNN failed to recover the lost performance confirming our intuition that mesh features provide additional information not easily derived from simple (stereo) camera features.

\begin{table}[]
  \centering
  \caption{Depth Error Correction Performance}
  
  \label{tab:error_metrics}
  \begin{tabular}{l|cccc}
    \toprule
    \textbf{Model}  & \textsc{rmse} & $\delta_1$ & $\delta_2$ & $\delta_3$ \\
    \midrule
    \borg~\cite{tanner2016bor} & 0.051 & 0.799 & 0.844 & 0.870 \\
    Our \ac{CNN} approach & \textbf{0.046} & \textbf{0.815} & \textbf{0.874} & \textbf{0.905} \\
    \bottomrule
  \end{tabular}
\end{table}

\begin{table}[tb!]
  \centering
  \caption{Fine-tuning with Reduced Feature Inputs}
  
  \label{tab:ablation_summary}
  \begin{tabular}{l|cccc}
    \toprule
    \textbf{Model} & $\delta_1$ \\
    \midrule
    \ac{CNN} with all features & 0.815 \\
    No area, surface-to-camera angle, edge ratios & 0.806 \\
    No area, surface-to-camera angle, edge ratios, fine-tuned & 0.807 \\
    \bottomrule
  \end{tabular}
  \vspace{-4mm}
\end{table}

\section{Conclusions}

In this paper we present a supervised learning technique for training a neural network to detect and correct error-prone regions of a dense reconstruction.
Using a \ac{GLSL} pipeline, we placed a virtual camera at various positions and orientations throughout the reconstruction to generate both input feature data and, by comparing the laser and camera reconstruction features, ground-truth data to train the network. We demonstrated that the error predictions of our model can  be used to make corrections to a mesh projected into an inverse-depth image and provide quantitative evaluation.

\ignore{
Trained on the reconstructions created in \cite{tanner_denser_2016} for KITTI-VO sequences 00, 05, and 06, our network's performance was quantitatively analysed using leave-one-out cross-validation.
Qualitatively, the network consistently recognised and highlighted the similar problem areas as those found in the ground-truth data.
We leave it to future work to use these network predictions to either remove or fix the error-prone reconstruction regions.
}

\section*{Acknowledgment}
The authors would like to acknowledge the support of the UK’s Engineering and Physical Sciences Research Council (EPSRC) through the Centre for Doctoral Training in Autonomous Intelligent Machines and Systems (AIMS) Programme Grant EP/L015897/1, Programme Grant EP/M019918/1, and the Doctoral Training Award (DTA).
Additionally, the donation from Nvidia of the Titan Xp GPU used in this work is also gratefully acknowledged.

\bibliographystyle{IEEEtran}
\bibliography{references}
~ 

\end{document}